\documentclass[twoside]{article}

\usepackage{graphicx}

\usepackage{CJKutf8}
\usepackage{svg}
\usepackage{acro}
\usepackage{amsmath}
\usepackage{amssymb}
\usepackage{xcolor}
\usepackage{algorithm}
\usepackage{algorithmic}
\usepackage[preprint]{aistats2026}
\usepackage[round]{natbib}

\begin{document}

% \runningtitle{Dual-Enhancement Product Bundling: Bridging Interactive Graph and Large Language Model}


\twocolumn[

\aistatstitle{Dual-Enhancement Product Bundling: Bridging Interactive Graph and Large Language Model}

\aistatsauthor{ 
  Zhe Huang \And Peng Wang \And Yan Zheng \And 
  Sen Song \And Longjun Cai 
}

\aistatsaddress{ 
  BUPT \\ Beijing, China \And 
  Beijing Wispirit Technology \\ Beijing, China \And 
  BUPT \\ Beijing, China \And 
  THU \\ Beijing, China \And  
  Beijing Wispirit Technology \\ Beijing, China 
}
]

\begin{abstract}

% Product bundling boosts e-commerce revenue by recommending complementary item combinations. However, due to the significant gap between linguistic processing objectives in \acp{LLM} and combinatorial optimization requirements in product bundling tasks, existing methods face two critical challenges: (1) graph-based approaches (e.g., LightGCN) struggle with cold-start items due to dependency on historical interactions, and (2) LLMs lack inherent capability to model graph-structured product relations. To bridge this gap, we propose a Dual-Enhancement method that synergistically integrates user-item graph learning and LLM-based semantic understanding for product bundling.

% Our method introduces two key innovations: (1) \ac{DCBM}: Aligns domain-specific product entities with LLM tokenization to resolve semantic ambiguity, enabling precise knowledge transfer from pretrained LLMs. (2) \acl{GTNM}: Translates graph-structured relationships into natural language prompts, allowing LLMs to natively comprehend combinatorial constraints.
% We further align LightGCN-learned graph features with LLM embeddings via an attentional projector, ensuring multimodal collaboration. Experiments on three benchmarks (POG, POG\_dense, Steam) demonstrate 6.3\%–26.5\% improvements in HitRatio@1 over state-of-the-art baselines, with 100\% ValidRatio in generated bundles. Our code is publicly available at [URL].

Product bundling boosts e-commerce revenue by recommending complementary item combinations. However, existing methods face two critical challenges: (1) collaborative filtering approaches struggle with cold-start items owing to dependency on historical interactions, and (2) LLMs lack inherent capability to model interactive graph directly. To bridge this gap, we propose a dual-enhancement method that integrates interactive graph learning and LLM-based semantic understanding for product bundling. Our method introduces a graph-to-text paradigm, which leverages a Dynamic Concept Binding Mechanism (DCBM) to translate graph structures into natural language prompts. The DCBM plays a critical role in aligning domain-specific entities with LLM tokenization, enabling effective comprehension of combinatorial constraints. Experiments on three benchmarks (POG, POG\_dense, Steam) demonstrate 6.3\%–26.5\% improvements over state-of-the-art baselines.
% Our method introduces a Dynamic Concept Binding Mechanism (DCBM) for aligning domain-specific entities with LLM tokenization, and 
% Our method introduces a graph-to-text paradigm to translate graph structures into natural language prompts for LLM comprehension of combinatorial constraints, a Dynamic Concept Binding Mechanism (DCBM) used to align domain-specific entities with LLM tokenization. 

\end{abstract}

\section{Introduction}

% ================================================================
% =====介绍背景
% ================================================================

\begin{figure}[h]
\vspace{.3in}
\centerline{\includegraphics[width=0.85\linewidth]{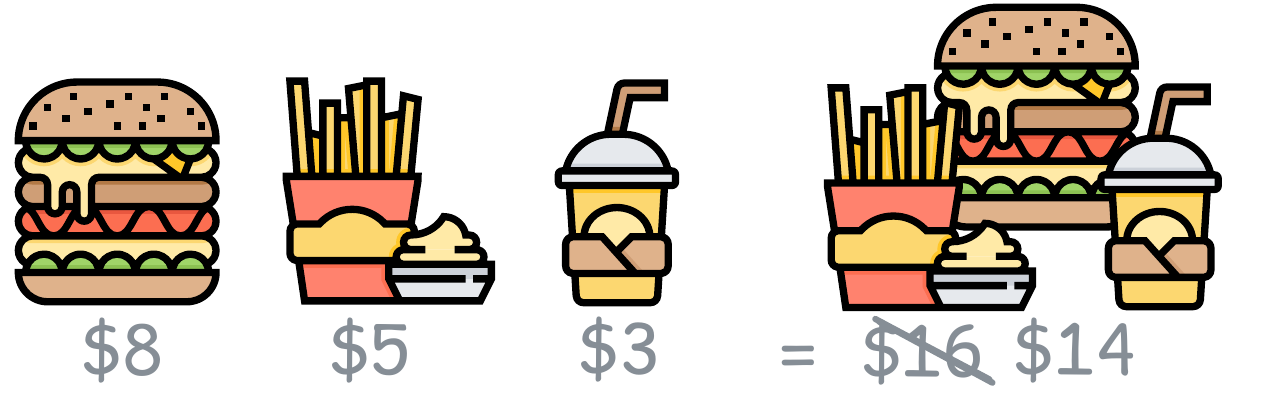}}
\vspace{.3in}
\caption{An example of product bundling.}
\label{fig:bundle}
\end{figure}

% Product bundling tasks strategically combine multiple items based on user behavioral preferences and product information to enhance purchase intention and transaction efficiency \citep{ProductBundlingInEcommerce}. This approach fundamentally differs from conventional sequential recommendation systems, which may recommend homogeneous products, as product bundles typically generate synergistic value. For instance, a product bundle containing smart speakers and smart home hubs enables users to conveniently control connected devices (e.g., smart lights and curtains) through voice commands, significantly improving the convenience and user experience of smart home ecosystems. This synergy, compared to sequential recommendations, enhances the attractiveness of bundled offerings by generating greater collaborative value.

Product bundling strategically combines complementary items based on user preferences and product information to boost purchase intention and efficiency \citep{ProductBundlingInEcommerce}. Unlike conventional sequential recommendations that suggest similar items, product bundling task creates synergistic solutions. 
For example, a bundle pairing a hamburger with fries and a cold drink not only offers a satisfying meal combination but also provides cost savings compared to purchasing each item separately, as shown in Figure~\ref{fig:bundle}.  The complementary value makes bundled recommendations significantly more appealing than standalone item suggestions.

% Our work draws inspiration from existing research in bundle recommendation systems \citep{BGCN,BunCa,CrossCBR,CoHeat,EBRec}, particularly in representation learning of product items.

\begin{figure}[h]
% \vspace{.3in}
\centerline{\includegraphics[width=1\linewidth]{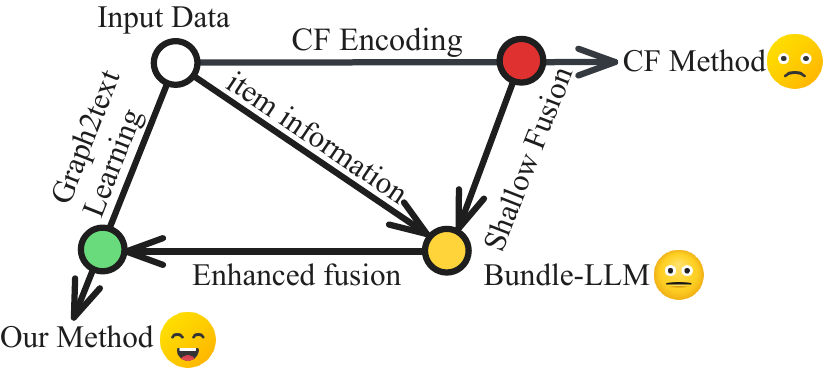}}
\vspace{.3in}
\caption{Different methods illustration on product bundling tasks.}
\label{fig:motivation}
\end{figure}

Although the Collaborative Filtering(CF) method has been widely used in recommendation systems and has demonstrated effectiveness \citep{CLHE}, it still faces significant challenges in handling cold-start items: newly introduced products lacking established interaction patterns, as visualized in Figure~\ref{fig:motivation}. This limitation is particularly evident, as the performance of CF method is heavily dependent on historical interaction data \citep{CoHeat}. 

% Research on domain-specific LLMs has also gradually emerged by combining professional domain knowledge and common-sense knowledge, can better understand and process text data in specific domains. For example, in the medical field, domain-specific LLMs can help doctors with disease diagnosis, drug recommendations, and medical record analysis; in the financial field, these models can be used for risk assessment, investment decisions, and market predictions, etc. The emergence of domain-specific LLMs not only improves the performance of models in professional domains but also provides strong support for product bundling tasks. 

Recent breakthroughs in LLM have revealed remarkable capabilities in semantic understanding and cross-domain knowledge transfer.
% , suggesting new possibilities for addressing these fundamental limitations. 
In view of this, recent attempts have leveraged LLMs to enhance CF-based method in bundle construction \citep{Bundle_LLM}. However, LLMs primarily process textual semantics and general domain knowledge, inherently lacking capability to comprehend graph-structured information in product bundling tasks. 
% 添加引文

% ================================================================
% =====介绍我们的方法
% ================================================================

To bridge this gap, we propose a dual-enhancement method based on interactive graph learning and LLM for product bundling. 
By leveraging the rich semantic understanding and reasoning capabilities of LLMs, we aim to enhance the overall effectiveness of relational modeling.
Specifically, the proposed method is a two-stage fine-tuning scheme for large language models. In the first stage, we use a dynamic concept-binding mechanism that maps item textual descriptions to domain-specific vocabularies, thereby bridging the gap between the pretraining knowledge of LLMs and product domain concepts. Further, we introduce a novel graph-to-text paradigm that converts interaction information into structured textual descriptions. The second stage focuses on product bundling task optimization, where LightGCN features are integrated into item representations, serving as input embedding to the large language model \citep{LightGCN}.
% In the first phase, we construct Task-Specific Adaptation Finetuning data to enable LLM adaptation to product bundling tasks by learning product item concepts and product-user-bundle relationships. 
% Our key innovation in this phase is the \ac{DCBM}, which addresses the lexical gap between the world knowledge acquired during LLM pretraining and domain-specific product knowledge in recommendation systems. This method establishes domain-specific tokenization concepts for product entities, effectively resolving semantic ambiguity in compound terms while constructing a semantic-decoupled embedding space to prevent conceptual confusion.Furthermore, we develop a \ac{GTNM} approach for relational data modeling. Unlike conventional multimodal LLMs that process extracted features through modality concatenation, our method innovatively converts graph structures into textual descriptions. Through this Task-Specific Adaptation Finetuning phase, we establish fundamental capabilities for \ac{LLM} to comprehend and process product items.
% In the second phase, we prepare task-specific data for product bundling by reformulating the retrieval task as a generation task to enhance LLM adaptation. 
% During this stage, we introduce a lightweight projector module that aligns LightGCN features, with the LLM's embedding space. This design further strengthens the model's understanding of relational information while maintaining compatibility with existing recommendation system components.

\begin{figure*}[h]
% \vspace{.3in}
\centerline{\includegraphics[width=1\linewidth]{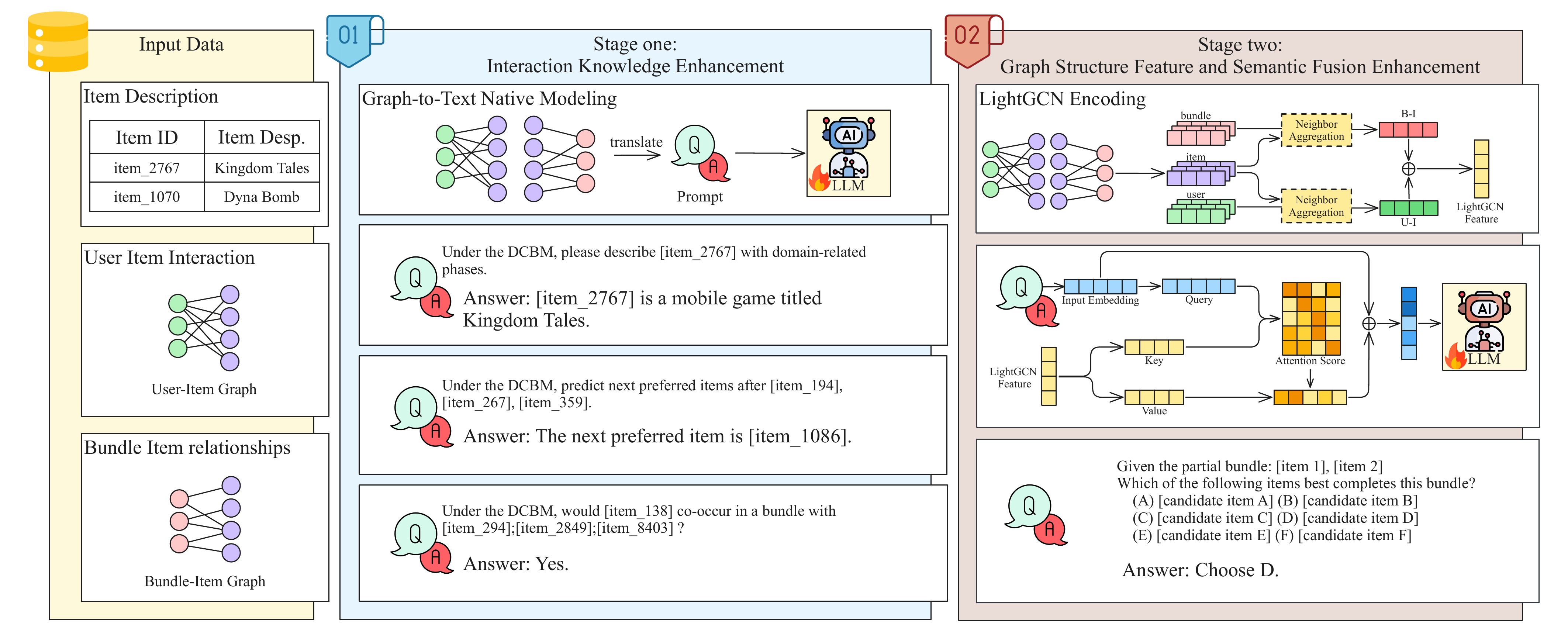}}
\vspace{.3in}
\caption{The dual-enhancement framework for product bundling.}
\label{fig:architecture}
\end{figure*}

% \includesvg[width=\textwidth]{figure/arch.svg}
% \begin{figure*}[t]
%     \centering
%     \includegraphics[width=0.9\linewidth]{figure/architecture.png}
%     \caption {A minimal working example to demonstrate how to place
%         two images side-by-side.}
%         \label{fig:architecture}
% \end{figure*}

We summarize the key contributions of this work as follows:
\begin{itemize}
    \item \textbf{Dual-Enhancement Framework}: We propose a novel dual-enhancement framework with two-stage fine-tuning.

    \item \textbf{Dynamic Concept Binding Mechanism}: An innovative adaptation method that enabling simultaneous preservation of LLMs' world knowledge and precise product entity binding.
    
    \item \textbf{Graph-to-text Native Modeling}: A graph-awared conversion paradigm transforms user-item interaction into textual descriptions, effectively alleviating information loss in cross-modal alignment.
  \end{itemize}
\section{Related Work}

% \subsection{Bundle Construction and Recommendation}

In the domain of bundle recommendation, several methods leverage pre-trained CF models to capture user-bundle interactions. \citet{ma2024leveraging} propose a framework that employs pre-trained CF models combined with content encoders to addressing the sparsity issue in user-bundle interactions effectively. \citet{chen2019matching} introduce an attention-based approach that aggregates item embeddings to represent bundles. \citet{li2023auto} formulate bundle recommendation as a link prediction task on graphs, designing compact spectral graph filters to capture user-bundle and bundle-item relationships. \citet{chang2020bundle} propose a heterogeneous graph framework that unifies relationships at both item and bundle levels, performing embedding propagation and exploring user preferences through hard negative sampling.

% Meanwhile, some approaches enhance recommendations through contrastive learning techniques. \citet{CrossCBR} develop cross-view contrastive learning to simultaneously leverage user-bundle and user-item interactions, learning view-specific representations to capture user preferences. \citet{ma2024multicbr} extend this concept by incorporating multi-view information through contrastive learning to obtain more robust representations. Additionally, \citet{bai2019personalized} model bundle list recommendation as a sequence generation problem, designing feature-aware softmax functions and specialized generation strategies to produce high-quality and diverse bundle lists. \citet{sun2024survey} provide a comprehensive survey of bundle recommendation methods, categorizing them into discriminative and generative paradigms while systematically analyzing their respective representation learning and interaction modeling approaches, offering valuable references for the field.

\subsection{Graph Convolutional Network in Product Bundling}

In recent years, the graph convolutional networks (GCNs) has seen rapid development, demonstrating strong capabilities for relationship modeling. Recent research has continuously explored the potential of GCN, giving rise to various improved and optimized models.

Among the many variants of GCN, LightGCN stands out with its unique advantages and has become a mainstream method for relationship modeling \citep{LightGCN}. The model structure of LightGCN is capable of effectively capturing high-order connection information and performing exceptionally well in handling complex graph structure data. This characteristic has made it widely adopted in research and applications such as \citet{BunCa,CrossCBR,CoHeat,EBRec,CIRP}.

 % Modern bundle recommendation systems increasingly adopt GCN-based approaches that extend traditional collaborative filtering principles to graph domains, leveraging architectures such as LightGCN . These methods typically construct heterogeneous graphs that incorporate item, user, and bundle nodes, with representation learning achieved through graph propagation mechanisms. 

% In the LightGCN model, the core mechanism lies in the propagation of feature information of product terms along the edges of the graph. As the number of network layers increases, each node can gradually integrate information from increasingly distant neighbor nodes, thus achieving multihop fusion. In this way, the model can better understand the relationships between nodes, thereby improving the accuracy and effectiveness of relationship modeling.

\subsection{LLMs in Bundle Construction}

Recently, there are remarkable breakthroughs in the development of LLMs, fueled by increasingly computing power and abundant data resources. The advancements in LLMs have inspired exploration of their potential applications in product bundling domains. As is known to all, conventional product bundling frameworks often rely on users' historical behavioral data and manually curated bundles, lacking in-depth understanding of individual product items and their interrelationships. The emergence of LLMs offers novel and strong solutions to address these limitations \citep{llminBG,Bundle_LLM,BRUCE}. By integrating LLMs with product bundling domains, we can leverage their capabilities to analyze product descriptions and user-item interactions, thereby more accurately capturing user interests and needs to achieve personalized and precise product bundling.
\section{Methodology}
In product bundling task, the domain-specific data construct an item interactive graph where products act as nodes, with latent connections established through user-item interactions and bundle-item affiliations.
To bridge the gap between LLMs and product bundling
tasks, we propose a \textbf{D}ual-enhancement \textbf{P}roduct \textbf{B}undling framework to finetune LLMs, named DPB-LLM. As shown in Figure \ref{fig:architecture}, DPB-LLM contains two-stage training objectives.
The corresponding pseudo-code, as illustrated in Algorithm~\ref{all:alogrithm}, show that our method presents a systematic approach to enhance product bundling.
\begin{algorithm*}[h]
    \caption{Dual-enhancement algorithm for product bundling.}
    \label{all:alogrithm}
    \begin{algorithmic}[1]
      \STATE \textbf{Input:} a base LLM $\mathcal{M}$, a set of items $\mathcal{I}=\{i_1,i_2,...,i_n\}$, the description of the items $\text{Desp}_{\mathcal{I}}=\{\text{Desp}_1,\text{Desp}_2,...,\text{Desp}_n\}$, the item sets interacted with by users $\mathcal{E}_u=\{\mathcal{E}_u^1,\mathcal{E}_u^2,...,\mathcal{E}_u^m\}$, the items sets in bundles $\mathcal{E}_b=\{\mathcal{E}_b^1,\mathcal{E}_b^2,...,\mathcal{E}_b^p\}$
        \FOR{$t=1$ to $n$}
            \STATE The tokenizer of $\mathcal{M}_0$ split the textual description $\text{desp}_t$ of item $i_t$ into tokens, which are then processed by $\mathcal{M}_0$'s embedding layer to derive the final embedding $E_t$, as formalized in Eq.~\eqref{eq:item_ids} and Eq.~\eqref{eq:item_emb}
            \STATE Build instruction dataset $\mathcal{D}_\text{node}=\{(\text{Item}_\text{desp}^t , \text{Item}_\text{node}^t)|t \in \mathcal{I}\}$, w.r.t. Eq.~\eqref{eq:label_align}
        \ENDFOR
        \FOR{$t=1$ to $m$}
            \FOR{$k \in \mathcal{E}_u^t$}
                \IF{index($k$) < len($\mathcal{E}_u^t$)//2}
                    \STATE add $k$ to $\mathcal{H}(t)$
                \ENDIF
            \ENDFOR
            \STATE Build instruction dataset $\mathcal{D}_\text{ui}=\{(\mathcal{E}_u^t \setminus \mathcal{H}(t), \mathcal{H}(t))|t \in \mathcal{E}_u\}$, w.r.t. Eq.~\eqref{eq:ui_align}.
        \ENDFOR
        \FOR{$t=1$ to $p$}
        
            \STATE Build instruction dataset $\mathcal{D}_\text{bi}=\{(True, \mathcal{E}_b^t)|t \in \mathcal{E}_b\}$, w.r.t. Eq.~\eqref{eq:bi_align}
            
            \STATE Sample a random item $j \gets \mathcal{I} \setminus \mathcal{E}_b$ 
            \STATE randomly selecting an item $k \in \mathcal{E}_b$ and replacing it with $j$ to obtained new set $\mathcal{B}'$
            \STATE Build instruction dataset $\mathcal{D}_\text{bi}=\{(False, \mathcal{B}')|t \in \mathcal{E}_b\}$, w.r.t. Eq.~\eqref{eq:bi_align}
        \ENDFOR
        \STATE Apply LoRA fine-tuning to $\mathcal{M}_0$ on dataset $\mathcal{D}'=\{\mathcal{D}_\text{node},\mathcal{D}_\text{ui},\mathcal{D}_\text{bi}\}$, yielding $\mathcal{M}_{\theta'}$
        \STATE train GCN model for user-item and bundle-item relational graph

        \FOR{$t=1$ to $p$}
            \FOR{$k \in \mathcal{E}_b^t$}
                \IF{index($k$) < len($\mathcal{E}_u^t$)//2}
                    \STATE add $k$ to $\mathcal{H}(t)$
                \ENDIF
            \ENDFOR
            
            \FOR{$k \in \mathcal{E}_b^t \setminus\mathcal{H}(t)$}
                \STATE Build instruction dataset $\mathcal{D}_\text{bp}=\{(\mathcal{E}_u^t, \mathcal{H}(t))|t \in \mathcal{E}_u\}$
                % , w.r.t. Eq.~\eqref{eq:label_align}
            \ENDFOR
            
        \ENDFOR
        \STATE On the $\mathcal{D}_\text{bp}$ dataset, optimize the model $\mathcal{M}_{\theta'}$ with parameters $\theta_\text{pb}$ to obtain DPB-LLM $\mathcal{M}_{\theta_\text{pb}}$

    \end{algorithmic}
    
\end{algorithm*}

% The quality of product representations fundamentally depends on three key factors: (1) the item's intrinsic informational attributes, (2) user-item interactions, and (3) item-bundle affiliations. This requires the simultaneous capture of reliable relational information and semantic understanding within product ecosystems. 

\subsection{Stage one: Interaction Knowledge Enhancement} 
% As shown in Figure \ref{fig:architecture}, 
In this stage, we translate the graph into natural language knowledge to finetune LLM, which aims to enhance LLMs' adaptability before undertaking product bundling tasks. 

%%%%%%%%% Item Desp.
% 

\subsubsection{Dynamic Concept Binding Mechanism}

\begin{figure}[t]
% \vspace{.3in}
    \centerline{\includegraphics[width=0.9\linewidth]{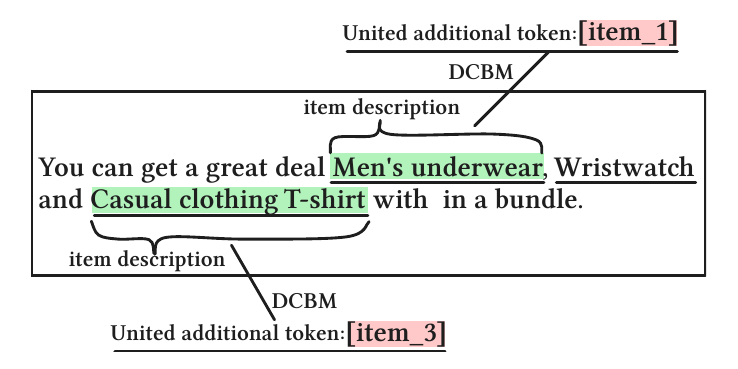}}
    \caption {Illustration of dynamic concept binding mechanism.}
        \label{fig:item_desp}
\vspace{.3in}
\end{figure}

For item representation, our proposed DCBM creates domain-specific tokenized concepts, where each item acts as a united additional token added to LLM's vocabulary. As shown in Figure~\ref{fig:item_desp}, we align these items with textual descriptions of products to address the lexical gap between the world knowledge acquired during LLM pretraining and domain-specific product knowledge in recommendation systems. This method effectively resolving semantic ambiguity in compound terms to prevent conceptual confusion.
We utilize item descriptions to initialize these items nodes $E_{i} \in \mathbb{R}^d$. 

As shown in Equation~\ref{eq:item_ids}, the item descriptions will be tokenized.
\begin{equation}
  \label{eq:item_ids}
  \text{Tokens}_i = \text{LLM}_\text{Tokenizer}(\text{desp}_i),
\end{equation}

where $\text{Tokens}_i$ is the token IDs of item $i$ description, $\text{desp}_i$ is the textual description of item $i$, and $\text{LLM}_\text{Tokenizer}(\cdot)$ denotes the LLM tokenizer that splits the description strings into $K$ sub-word token strings. Then, we can obtain the embedding $E_{i}$ of item $i$ as Equation~\ref{eq:item_emb}.

\begin{equation}
  \label{eq:item_emb}
  E_{i} = \frac{1}{K}\sum_{k=1}^K \text{LLM}_{\text{Embedding}}\left(\text{Tokens}_i^{(k)}\right),
\end{equation}

where $\text{LLM}_{\text{Embedding}}(\cdot)$ represents the embedding layer of the LLMs. After obtaining $E_{i}$, we design the interaction enhancement prompt template as figure \ref{fig:architecture}.
% \begin{quote}
% \textbf{Input:} Under the DCBM, please describe [$\text{Item}_\text{node\_id}$] with domain-related phases.\\
% \textbf{Output:} [$\text{Desp}_\text{node\_id}$]
% \end{quote}
Here item IDs corresponds to a additional token with the learnable node $E_{i}$, while item descriptions contains the actual textual description, as shown in Figure~\ref{fig:item_desp}. The mechanism optimizes via Equation~\ref{eq:label_align}.

\begin{equation}
  \label{eq:label_align}
  \mathcal{L}_{\text{DCBM}} = -\sum_{i=1}^N \log P(\text{Desp}_i | \text{Item}_i; \theta),
\end{equation}

where $\theta$ denotes the LoRA parameters of the LLM. The objective is to minimize the negative log-likelihood of the model generating the textual description given the item node, effectively aligning the node with the product's textual information.

\subsubsection{Graph-to-Text Native Modeling}
Based on DCBM, we translate user-item and bundle-item graph into QA pair separately through our proposed graph-to-text native modeling method, enabling LLMs to comprehend complex relationships. This approach fully leverages LLMs' native text comprehension capabilities to enhance semantic modeling of structured relationships, effectively alleviating information loss in cross-modal feature enhancement.

To model user-item interactions, we define the enhancement objective as:

\begin{equation}
  \label{eq:ui_align}
  \mathcal{L}_{\text{user}} = -\sum_{u \in \mathcal{U}} \log P(S_u \setminus \mathcal{H}_u| \mathcal{H}_u; \theta),
\end{equation}

where $\mathcal{U}$ denotes the user set, $S_u = \{i | (u,i) \in \mathcal{E}_u\}$ represents the complete interactive item sequence of user $u$. Let $\mathcal{H}_u \subseteq S_u$ denote a partial interaction sequence sampled from $S_u$, representing the observed history of user $u$. The conditional probability $P(S_u \setminus \mathcal{H}_u \mid \mathcal{H}_u; \theta)$ models the likelihood of the user's future interactions (i.e., $S_u \setminus \mathcal{H}_u$) given the observed history $\mathcal{H}_u$, where $\theta$ represents the model parameters. We design user-item relational enhancement templates for prompting LLM as figure \ref{fig:architecture}.

% \begin{quote}
% \textbf{Input:} Under the DCBM, predict next preferred items after [$\text{Item}_x$];[$\text{Item}_y$]... .\\
% \textbf{Output:} [$\text{Item}_{seq\_1}$];[$\text{Item}_2$];[$\text{Item}_3$]... .
% \end{quote}

Following the same principle, we formulate the bundle-item relationship enhancement objective as:

\begin{equation}
  \label{eq:bi_align}
    \mathcal{L}_{\text{bundle}} = -\sum_{b\in\mathcal{B}} \log P(b \in \mathcal{B}^* | \mathcal{C}(b); \theta),
\end{equation}

where $\mathcal{B}^*$ denotes the ground-truth bundle collection, $\mathcal{C}(b)$ represents the candidate item set for bundle construction, and the conditional probability $P(b \in \mathcal{B}^* | \mathcal{C}(b); \theta)$ measures how likely the item set $\mathcal{C}(b)$ forms a valid bundle.
We design bundle-item relational enhancement templates as figure \ref{fig:architecture}.

% \begin{quote}
%     \textbf{Input:} 
    
%     Under the DCBM, would [$\text{Item}_x$] co-occur in a bundle with [$\text{Item}_1$];[$\text{Item}_2$];[$\text{Item}_3$]... ? \\
%     \textbf{Output:} [yes/no].
% \end{quote}

\subsection{Stage two: Graph Structure Feature and Semantic Fusion Enhancement}
% , as shown in Figure \ref{fig:architecture}b
The second stage focuses on cross-modal feature fusion. Here, we preprocess training data to conduct product bundling as an instruction-finetuned text generation task. For structure features, we employ the attentional projector to align LightGCN features with LLM text representations in the latent space, preserving original graph modeling advantages while enhancing semantic understanding through this lightweight design.

\subsubsection{Fusion Method}
To integrate the LightGCN features with the LLM semantics, we employ a cross-attention mechanism to fuse the item representations $\mathbf{I}_b$ and $\mathbf{I}_u$ with the text embeddings.

First, we map the two item representations into the text embedding dimension through a two-layer MLP:
\begin{equation}
\mathbf{I}_{\text{multi}} = \left[ \mathbf{I}_b \| \mathbf{I}_u \right] \in \mathbb{R}^{B \times M \times 2d},
\label{eq:multi_modal}
\end{equation}
\begin{equation}
\mathbf{I}_{\text{mapped}} = \text{GELU}\left(\mathbf{I}_{\text{multi}} \mathbf{W}_1 + \mathbf{b}_1\right) \mathbf{W}_2 + \mathbf{b}_2,
\label{eq:mlp_combined}
\end{equation}
where $\mathbf{W}_1 \in \mathbb{R}^{2d \times d_{\text{hidden}}}$, $\mathbf{W}_2 \in \mathbb{R}^{d_{\text{hidden}} \times d_h}$, $\mathbf{b}_1 \in \mathbb{R}^{d_{\text{hidden}}}$, and $\mathbf{b}_2 \in \mathbb{R}^{d_h}$ are learnable parameters.

For the cross-attention mechanism, we compute the attention output directly from the query, key, and value matrices:
\begin{equation}
\mathbf{Q} = \mathbf{T} \mathbf{W}_q, \quad \mathbf{K} = \mathbf{I}_{\text{mapped}} \mathbf{W}_k, \quad \mathbf{V} = \mathbf{I}_{\text{mapped}} \mathbf{W}_v,
\label{eq:qkv_cross}
\end{equation}
\begin{equation}
\mathbf{A}_{\text{output}} = \text{softmax}\left(\frac{\mathbf{Q} \mathbf{K}^\top}{\sqrt{d_h}}\right) \mathbf{V} \in \mathbb{R}^{B \times T \times d_h},
\label{eq:cross_attn_combined}
\end{equation}
where $\mathbf{W}_q, \mathbf{W}_k, \mathbf{W}_v \in \mathbb{R}^{d_h \times d_h}$ are learnable projection matrices.

Finally, the fused representation is obtained through projection, residual connection, and layer normalization:
\begin{equation}
\mathbf{I}_{\text{fusion}} = \text{LayerNorm}(\mathbf{T} + \mathbf{A}_{\text{output}} \mathbf{W}_{\text{out}}),
\label{eq:final_fusion}
\end{equation}
where $\mathbf{W}_{\text{out}} \in \mathbb{R}^{d_h \times d_h}$ is a learnable output projection matrix.

\subsubsection{Product Bundling Task Finetuning}

In this phase, we perform supervised fine-tuning on the LLMs to optimize the product bundling. We construct a specialized fine-tuning dataset where the integrate representations $\mathbf{I}_{\text{fusion}}$ from the previous cross-attention mechanism serve as input features to guide the model's understanding of item relationships and bundle composition patterns.

The training data is formulated as a multiple-choice question-answering task, where each instance consists of a partially specified bundle along with a set of candidate items. Specifically, given a target bundle $\mathcal{B} = \{i_1, i_2, ..., i_k\}$, we construct training examples by masking one item $i_j \in \mathcal{B}$ and presenting the remaining items $\mathcal{B}_{-j} = \mathcal{B} \setminus \{i_j\}$ as context. The model is then provided with a candidate set $\mathcal{C} = \{c_1, c_2, ..., c_n\}$ containing the ground truth item $i_j$ along with negative samples, and is required to select the correct completion item.

% \begin{quote}
% Given the partial bundle: [item descriptions from $\mathcal{B}_{-j}$]
% Which of the following items best completes this bundle?

% A) [candidate item $c_1$]

% B) [candidate item $c_2$]

% ...

% N) [candidate item $c_n$]

% \end{quote}

As shown in Figure \ref{fig:architecture}, the input prompt follows a structured template that incorporates both the contextual bundle information and the fused representations.
The LLM processes this structured input augmented with the cross-attention fused features $\mathbf{I}_{\text{fusion}}$ and generates a probability distribution over the candidate choices. The training objective employs cross-entropy loss to maximize the likelihood of the correct answer:

\begin{equation}
\mathcal{L}_{\text{SFT}} = -\sum_{i=1}^{|\mathcal{D}|} \log P(y_i | \mathcal{B}_{-j}^{(i)}, \mathcal{C}^{(i)}, \mathbf{I}_{\text{fusion}}^{(i)}),
\label{eq:sft_loss}
\end{equation}

where $\mathcal{D}$ represents the fine-tuning dataset, $y_i$ is the ground truth label for the $i$-th training instance, and $P(\cdot)$ denotes the probability distribution output by the LLM. This supervised learning approach enables the model to learn complex bundle completion patterns by leveraging both the semantic understanding from the pre-trained LLM and the collaborative filtering insights embedded in the fused representations.
\section{Experiment}
\subsection{Dataset}
\begin{table}
\caption{The statistics of the three datasets on two different domains.}
\label{tab:datasets}
\begin{center}
\begin{tabular}{lccc}

\textbf{Datasets} & \textbf{\#U} & \textbf{\#I} & \textbf{\#B} \\ \hline \\
        POG        &     17,449 & 48,676 & 20,000 \\
        POG\_dense &  2,311,431 & 31,217 & 29,686 \\
        steam      &     54,985 &  2,798 &    615 \\ 
\end{tabular}
\end{center}
\end{table}
We use three datasets from two domains to validate the effectiveness of proposed DPB-LLM, as shown in Table~\ref{tab:datasets}. Among them, POG is about fashion outfit combinations \citep{POG}, POG\_dense is a variant of POG with more dense user feedback, and steam \citep{steam_ref_R,steam_ref_SR,steam_ref_BR} is about game bundle sales. The data processing follows general settings, and we randomly divide all bundle in each dataset into training set, validation set, and test set, with a ratio of 8:1:1. The validation dataset is used to evaluate performance and find the optimal hyperparameters.

% \subsection{Baseline}
% In order to evaluate the method we proposed, I selected the currently mainstream methods as comparisons.

% \begin{itemize}
%     \item \textbf{ICL} \citep{ICL} constructs the task by providing the context of similar questions and answers with the LLM backbones including Qwen2-7B, LLaMA2-7B and Deepseek-R1.
    
%     \item \textbf{MultiVAE} \citep{MultiVAE} employs the variational autoencoder framework to encode discrete item-user interactions into continuous latent embeddings for recommendation tasks.

%     \item \textbf{BiLSTM} \citep{BiLSTM} utilizes bidirectional Long Short-Term Memory networks to capture sequential dependencies in bundle sequences and learn comprehensive bundle representations.
    
%     \item \textbf{UHBR} \citep{UHBR} constructs hypergraph structures to model intra-bundle relationships and applies Graph Convolutional Networks to derive bundle embeddings.
    
%     \item \textbf{CLHE} \citep{CLHE} implements a hierarchical encoding architecture integrated with contrastive learning mechanisms to generate discriminative bundle representations.

%     \item \textbf{Bundle-MLLM} \citep{Bundle_LLM} integrates hybrid tokenization and multimodal fusion with progressive optimization to unify diverse data types, enhancing cross-domain adaptability.
    
% \end{itemize}

\subsection{Baselines}
We compare our proposed method against the following state-of-the-art baselines:

\begin{itemize}
    \item \textbf{ICL} \citep{ICL}: The ability of a language model to perform tasks by leveraging examples and context provided within the input, rather than relying solely on pre-trained knowledge.
    
    \item \textbf{MultiVAE} \citep{MultiVAE}: Variational autoencoder framework that encodes discrete item-user interactions into continuous latent embeddings.
    
    \item \textbf{BiLSTM} \citep{BiLSTM}: Bidirectional LSTM networks for capturing sequential dependencies and learning bundle representations.
    
    \item \textbf{UHBR} \citep{UHBR}: Hypergraph-based approach that models bundle relationships using Graph Convolutional Networks.
    
    \item \textbf{CLHE} \citep{CLHE}: Hierarchical encoding with contrastive learning for discriminative bundle representation.
    
    \item \textbf{Bundle-MLLM} \citep{Bundle_LLM}: Multimodal approach with hybrid tokenization and progressive optimization for cross-domain adaptability.
\end{itemize}
\subsection{Implementation Details}
This experiment uses the Llama-2-7b-Chat and Qwen2-7B-Instruct as the base LLM for fine-tuning with the LoRA method. The LoRA configuration sets the rank to 16 and the scaling factor alpha to 32, and performs low-rank adaptation on the modules of the model (including the k\_proj, v\_proj, q\_proj, o\_proj of the attention layer, the gate\_proj, up\_proj, down\_proj of the MLP layer, as well as the lm\_head and embed\_tokens). Then, 2 GPUs are used for BF16 mixed-precision training, with 10 training epochs, using a linear warm-up cosine learning rate schedule with a base learning rate of 8e-5, starting from a warm-up initial learning rate of 8e-6 and gradually increasing to the base rate during the warm-up phase, and batch size is 8.

\subsection{Evaluation Metrics}

We adopt HitRate@1 as the primary metric, measuring the model's accuracy in selecting the most suitable product from 10 candidates. Additionally, we introduce ValidRatio to assess response validity, quantifying the proportion of valid predictions among all generated responses.

\begin{equation}
    \label{eq:hitrate}
    \text{HitRate}@1 = \frac{1}{N} \sum_{i=1}^{N} \mathbb{I}(\hat{y}_i = y_i),
\end{equation}
\begin{equation}
    \label{eq:validratio}
    \text{ValidRatio} = \frac{1}{N} \sum_{i=1}^{N} \mathbb{I}(\text{valid}(r_i)),
\end{equation}

where $N$ denotes the total number of test samples, $\hat{y}_i$ and $y_i$ represent predicted and ground truth labels respectively, $r_i$ is the generated response, and $\mathbb{I}(\cdot)$ is the indicator function.

\subsection{Experimental Results}

\begin{table*}[h]
\caption{The results of DPB-LLM compared with baseline methods.}
\label{tab:results}
\begin{center}
\begin{tabular}{lcccccc}

\textbf{Dataset} & \multicolumn{2}{c}{\textbf{POG}} & \multicolumn{2}{c}{\textbf{POG\_dense}} & \multicolumn{2}{c}{\textbf{steam}} \\ \hline \\
\textbf{Model} & \textbf{HitRate@1} & \textbf{ValidRatio} & \textbf{HitRate@1} & \textbf{ValidRatio} & \textbf{HitRate@1} & \textbf{ValidRatio} \\ \hline \\
ICL$^\dagger$                   &         0.10  & 1.00 &         0.08  & 1.00 &         0.12  & 0.13 \\ 
ICL$^\ddagger$                  &         0.10  & 1.00 &         0.10  & 1.00 &         0.57  & 0.83 \\ 
ICL$^{\bowtie}$                   &         0.23  & 0.79 &         0.20  & 0.78 &         0.76  & 0.98 \\
MultiVAE                        &         0.28  & 1.00 &         0.45  & 1.00 &          -    &  -   \\
BiLSTM                          &         0.29  & 1.00 &         0.43  & 1.00 &          -    &  -   \\
UHBR                            &         0.25  & 1.00 &         0.59  & 1.00 &          -    &  -   \\
CLHE                            &         0.35  & 1.00 &         0.64  & 1.00 &          -    &  -   \\
Bundle-MLLM$^\dagger$           &         0.41  & 1.00 &         0.67  & 1.00 &          -    &  -   \\
\textbf{DPB-LLM}$^\dagger$      &         0.49  & 1.00 &         0.71  & 1.00 &         0.81  & 1.00 \\
\textbf{DPB-LLM}$^\ddagger$     & \textbf{0.52} & 1.00 & \textbf{0.72} & 1.00 & \textbf{0.83} & 1.00 \\
\end{tabular}
\end{center}
\end{table*}

Table~\ref{tab:results} presents a comprehensive comparison of DPB-LLM against baseline methods across the evaluated datasets, where ${\bowtie}$ denotes DeepSeek-R1, $\dagger$ denotes LLaMA2-7B, and $\ddagger$ denotes Qwen2-7B as backbones. 

Our purposed DPB-LLM consistently outperforms all baseline methods across three datasets and achieve improvements of 6.3\% to 26.5\% in HitRate@1 against state-of-the-art model, which demonstrates superior bundling capability. While Bundle-MLLM leverages multimodal information(text and image), our approach operates exclusively on textual features. To further validate our model's effectiveness in text-only scenarios, we specifically include the Steam dataset, which contains purely textual information.

Cross-dataset analysis reveals that Qwen2 exhibits superior performance compared to LLaMA2 on both Chinese and English datasets. For the ValidRatio evaluation metric, DPB-LLM achieves perfect 100\% effectiveness across all datasets. 
% Our fine-tuning methodology substantially improves the untrained LLaMA2 and Qwen2 models while preserving their instruction-following capabilities.

\subsection{Ablation Studies}
% We conducted ablation studies to evaluate the impact of different components on DPB-LLM, as shown in Table~\ref{tab:ablation}. Here, we present five different strategies to analyze their effects:
We conducted ablation studies to evaluate the impact of different components on DPB-LLM from three perspectives, as shown in Table~\ref{tab:ablation}.

\subsubsection{Component Integration Analysis}
We analyse the effectiveness of integrating LLM with graph-based CF method from three configurations. 
\textbf{PB-LLM} refers to single stage enhancement via product bundling finetuning. 
% by projecting LightGCN features and concatenating them with product descriptions, directly incorporating collaborative filtering information. 
\textbf{PB-LLM w/o GCN} remove the LightGCN feature fusion.
\textbf{DPB-LLM} achieves optimal performance through our proposed dual enhancement method with two-stage training, effectively combining textual understanding and structural knowledge.

\subsubsection{Cold-start Performance Analysis}
To evaluate performance when items lack sufficient interaction information, we examine three scenarios. \textbf{PB-LLM w/o GCN} relies solely on textual understanding, showing LLM's capability in cold-start situations. \textbf{DPB-LLM w/o GCN} first enhances the model with interaction knowledge then fine-tunes on product descriptions, maintaining decent performance even without GCN features during inference, demonstrating the robustness of our dual enhancement approach.

\subsubsection{Training Strategy Analysis}
We compare our sequential two-stage approach with joint training to validate stepwise strategy. \textbf{DPB-LLM JT} simultaneously optimizes both interaction knowledge enhancement and feature fusion enhancement. \textbf{DPB-LLM} employs our two-stage training that progressively builds understanding from textual comprehension to feature integration, avoiding interference issues and achieving superior performance through systematic capability development.

\begin{table*}[h]
\caption{Ablation studies related to DPB-LLM.}
\label{tab:ablation}
\begin{center}
\begin{tabular}{lcccccc}

\textbf{Dataset} & \multicolumn{2}{c}{\textbf{POG}} & \multicolumn{2}{c}{\textbf{POG\_dense}} & \multicolumn{2}{c}{\textbf{steam}} \\ \hline \\
\textbf{Model} & \textbf{HitRate@1} & \textbf{ValidRatio} & \textbf{HitRate@1} & \textbf{ValidRatio} & \textbf{HitRate@1} & \textbf{ValidRatio} \\ \hline \\
        % \textbf{LLaMA2-7B:} &  & \\
        PB-LLM  w/o GCN$^\dagger$  &         0.44   & 1.00 &         0.65  & 1.00&         0.75  & 1.00\\ 
        PB-LLM$^\dagger$           &         0.46   & 1.00 &         0.67  & 1.00&         0.76  & 1.00\\
        DPB-LLM w/o GCN$^\dagger$  &         0.48   & 1.00 &         0.70  & 1.00&         0.80  & 1.00\\
        DPB-LLM-JT$^\dagger$       &         0.43   & 1.00 &         0.65  & 1.00&         0.77  & 1.00\\
        DPB-LLM$^\dagger$          & \textbf{0.49}  & 1.00 & \textbf{0.71} & 1.00& \textbf{0.81} & 1.00\\
        \hline \\
        % \textbf{Qwen2-7B:} &  & \\
        PB-LLM  w/o GCN$^\ddagger$  &         0.44  & 1.00 &         0.66  & 1.00&         0.78  & 1.00\\ 
        PB-LLM$^\ddagger$           &         0.49  & 1.00 &         0.67  & 1.00&         0.79  & 1.00\\
        DPB-LLM w/o GCN$^\ddagger$  &         0.50  & 1.00 &         0.69  & 1.00&         0.81  & 1.00\\
        DPB-LLM-JT$^\ddagger$       &         0.42  & 1.00 &         0.67  & 1.00&         0.79  & 1.00\\
        DPB-LLM$^\ddagger$          & \textbf{0.52} & 1.00 & \textbf{0.72} & 1.00& \textbf{0.83} & 1.00\\
\end{tabular}
\end{center}
\end{table*}

\subsection{Impact of Different Factors}

\subsubsection{different scales of LLM parameters}
\begin{figure}
    % \vspace{.3in}
    \centerline{\includegraphics[width=0.8\linewidth]{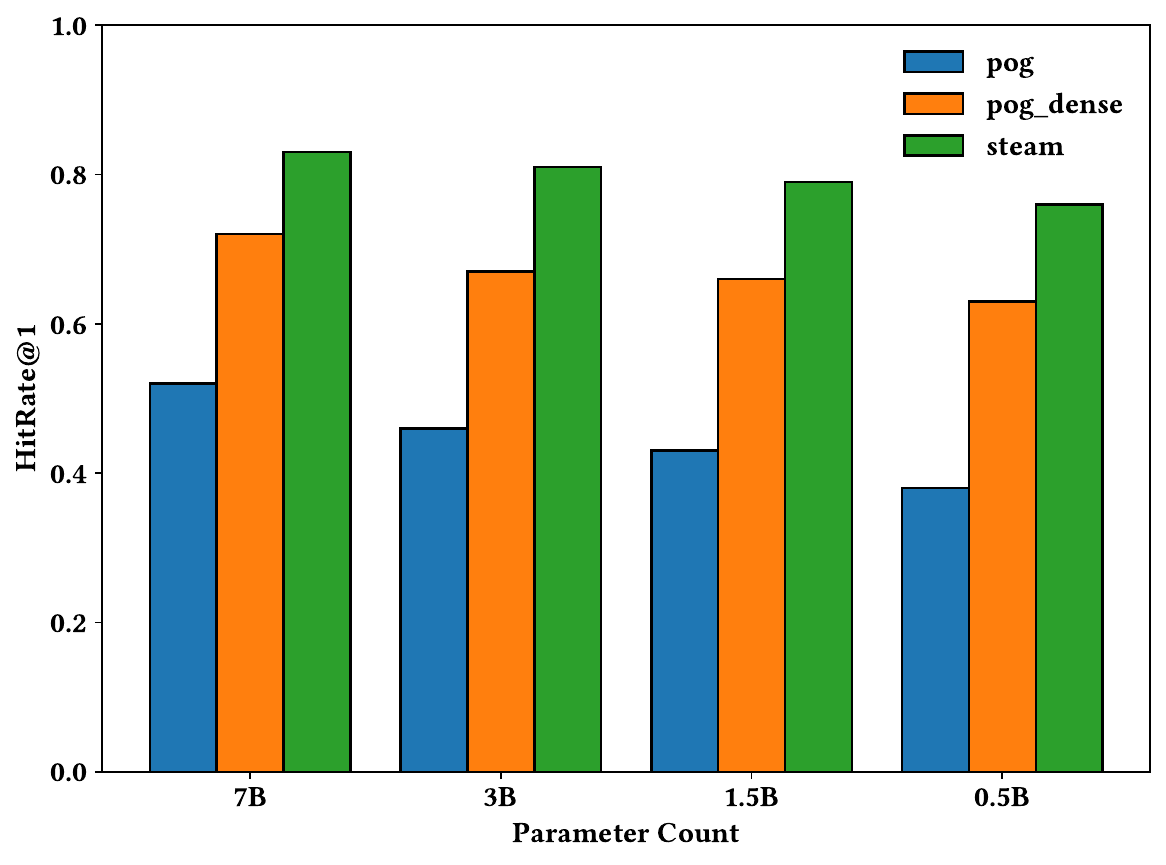}}
    \vspace{.3in}
    \caption{Comparison of different LLM parameter scales.}
    \label{fig:backbone}
\end{figure}
% We conducted comparative experiments on different scales of LLM (0.5B to 7B) based on Qwen2. As shown in the figure~\ref{fig:backbone}, it demonstrates consistent performance advantages for the steam dataset over pog and pog\_dense across all model scales. Meanwhile, all dataset exhibit gradual declines with smaller scales of LLM.

We conducted comparative experiments on different scales of LLM parameters (from 7B to 0.5B) based on Qwen2, as shown in the figure~\ref{fig:backbone}. The steam dataset maintains relatively stable performance across LLM scale reductions, while the pog\_dense and pog dataset displays accelerated performance decline. All datasets follow monotonic performance reduction aligned with fundamental scaling laws. The divergence in performance degradation highlights product bundling task dependence on model parameter scale in different datasets.

\begin{figure}
    % \vspace{.3in}
    \centerline{\includegraphics[width=\linewidth]{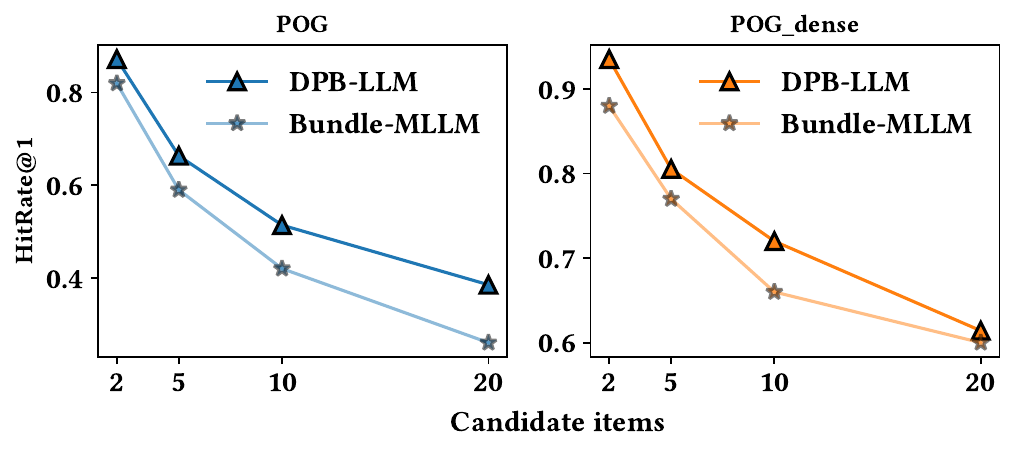}}
    \vspace{.3in}
    \caption{Comparison of different numbers of candidate items.}
    \label{fig:cans}
\end{figure}

\subsubsection{different numbers of candidate items}

We conducted experiments on the impact of the item candidate scale. As shown in the figure~\ref{fig:cans}, with an increase in the number of candidate items, the difficulty of the task also increases and the accuracy of the prediction decreases. However, DPB-LLM still ensures a stable accuracy rate. This proves the stability of our proposed method and highlights its reliable and effective application potential in complex scenarios.
\section{Conclusion and Future Work}
Traditional product bundling systems face significant limitations in simultaneously capturing comprehensive world knowledge and understanding user behavior patterns. This paper introduces DPB-LLM, a dual-enhanced product bundling framework that leverages interaction graph enhancement to inject collaborative filtering knowledge into large language models. Extensive experiments on three datasets demonstrate that our framework significantly outperforms state-of-the-art baseline models, achieving performance improvements ranging from 6.3\% to 26.5\%. These results validate the effectiveness of combining traditional recommendation systems' relational modeling capabilities with LLMs' world knowledge encoding strengths. 

Despite these promising results, several directions warrant future exploration. First, investigating more efficient knowledge injection methods could further unlock LLMs' potential in understanding product semantics and complex user-item relationships. Second, evaluating the framework across diverse datasets and real-world scenarios would better assess its generalization capability. Finally, optimizing the balance between model complexity and computational efficiency remains crucial for large-scale real-time recommendation deployment.

\bibliography{reference}

@inproceedings{CIRP,
	address = {Melbourne VIC Australia},
	title = {{CIRP}: {Cross}-{Item} {Relational} {Pre}-training for {Multimodal} {Product} {Bundling}},
	isbn = {979-8-4007-0686-8},
	shorttitle = {{CIRP}},
	url = {https://dl.acm.org/doi/10.1145/3664647.3681349},
	doi = {10.1145/3664647.3681349},
	language = {en},
	urldate = {2024-11-12},
	booktitle = {Proceedings of the 32nd {ACM} {International} {Conference} on {Multimedia}},
	publisher = {ACM},
	author = {Ma, Yunshan and He, Yingzhi and Zhong, Wenjun and Wang, Xiang and Zimmermann, Roger and Chua, Tat-Seng},
	month = oct,
	year = {2024},
	pages = {9641--9649},
}

@inproceedings{Bundle_LLM,
  title={Fine-tuning Multimodal Large Language Models for Product Bundling},
  author={Liu, Xiaohao and Wu, Jie and Tao, Zhulin and Ma, Yunshan and Wei, Yinwei and Chua, Tat-seng},
  booktitle={Proceedings of the 31st ACM SIGKDD Conference on Knowledge Discovery and Data Mining V. 1},
  pages={848--858},
  year={2025}
}

@misc{EBRec,
	title = {Enhancing {Item}-level {Bundle} {Representation} for {Bundle} {Recommendation}},
	shorttitle = {{EBRec}},
	url = {http://arxiv.org/abs/2311.16892},
	doi = {10.48550/arXiv.2311.16892},
	language = {en-US},
	urldate = {2024-10-30},
	publisher = {arXiv},
	author = {Du, Xiaoyu and Qian, Kun and Ma, Yunshan and Xiang, Xinguang},
	month = nov,
	year = {2023},
	note = {arXiv:2311.16892},
}

@misc{BunCa,
	title = {Bundle {Recommendation} with {Item}-level {Causation}-enhanced {Multi}-view {Learning}},
	shorttitle = {{BunCa}},
	url = {http://arxiv.org/abs/2408.08906},
	language = {en-US},
	urldate = {2024-11-04},
	publisher = {arXiv},
	author = {Nguyen, Huy-Son and Bui, Tuan-Nghia and Nguyen, Long-Hai and Manh-Hung, Hoang and Nguyen, Cam-Van Thi and Le, Hoang-Quynh and Le, Duc-Trong},
	month = aug,
	year = {2024},
	note = {arXiv:2408.08906 [cs]},
}

@inproceedings{Lightgcn,
	title = {Lightgcn: {Simplifying} and powering graph convolution network for recommendation},
	shorttitle = {{LightGCN}},
	language = {en},
	booktitle = {Proceedings of the 43rd {International} {ACM} {SIGIR} conference on research and development in {Information} {Retrieval}},
	author = {He, Xiangnan and Deng, Kuan and Wang, Xiang and Li, Yan and Zhang, Yongdong and Wang, Meng},
	year = {2020},
	pages = {639--648},
}

@inproceedings{CrossCBR,
	address = {Washington DC USA},
	title = {{CrossCBR}: {Cross}-view {Contrastive} {Learning} for {Bundle} {Recommendation}},
	isbn = {978-1-4503-9385-0},
	shorttitle = {{CrossCBR}},
	url = {https://dl.acm.org/doi/10.1145/3534678.3539229},
	doi = {10.1145/3534678.3539229},
	language = {en},
	urldate = {2024-11-13},
	booktitle = {Proceedings of the 28th {ACM} {SIGKDD} {Conference} on {Knowledge} {Discovery} and {Data} {Mining}},
	publisher = {ACM},
	author = {Ma, Yunshan and He, Yingzhi and Zhang, An and Wang, Xiang and Chua, Tat-Seng},
	month = aug,
	year = {2022},
	pages = {1233--1241},
}

@misc{CoHeat,
	title = {Cold-start {Bundle} {Recommendation} via {Popularity}-based {Coalescence} and {Curriculum} {Heating}},
	shorttitle = {{CoHeat}},
	url = {http://arxiv.org/abs/2310.03813},
	doi = {10.48550/arXiv.2310.03813},
	language = {en-US},
	urldate = {2024-10-30},
	publisher = {arXiv},
	author = {Jeon, Hyunsik and Lee, Jong-eun and Yun, Jeongin and Kang, U.},
	month = mar,
	year = {2024},
	note = {arXiv:2310.03813},
}

@article{ProductBundlingInEcommerce,
author = {Sun, Zhu and Feng, Kaidong and Yang, Jie and Fang, Hui and Qu, Xinghua and Ong, Yew-Soon and Liu, Wenyuan},
title = {Revisiting Bundle Recommendation for Intent-aware Product Bundling},
year = {2024},
issue_date = {September 2024},
publisher = {Association for Computing Machinery},
address = {New York, NY, USA},
volume = {2},
number = {3},
url = {https://doi.org/10.1145/3652865},
doi = {10.1145/3652865},
journal = {ACM Trans. Recomm. Syst.},
month = jun,
articleno = {24},
numpages = {34}
}

@inproceedings{POG,
  title={POG: personalized outfit generation for fashion recommendation at Alibaba iFashion},
  author={Chen, Wen and Huang, Pipei and Xu, Jiaming and Guo, Xin and Guo, Cheng and Sun, Fei and Li, Chao and Pfadler, Andreas and Zhao, Huan and Zhao, Binqiang},
  booktitle={Proceedings of the 25th ACM SIGKDD international conference on knowledge discovery \& data mining},
  pages={2662--2670},
  year={2019}
}

@inproceedings{steam_ref_R,
	title = {Item recommendation on monotonic behavior chains},
	shorttitle = {{steamDS}},
	booktitle = {Proceedings of the 12th {ACM} conference on recommender systems},
	author = {Wan, Mengting and McAuley, Julian},
	year = {2018},
	pages = {86--94},
}

@inproceedings{steam_ref_SR,
	title = {Self-attentive sequential recommendation},
	shorttitle = {{steamDS}},
	booktitle = {2018 {IEEE} international conference on data mining ({ICDM})},
	publisher = {IEEE},
	author = {Kang, Wang-Cheng and McAuley, Julian},
	year = {2018},
	pages = {197--206},
}

@inproceedings{steam_ref_BR,
	title = {Generating and personalizing bundle recommendations on steam},
	shorttitle = {{steamDS}},
	booktitle = {Proceedings of the 40th international {ACM} {SIGIR} conference on research and development in information retrieval},
	author = {Pathak, Apurva and Gupta, Kshitiz and McAuley, Julian},
	year = {2017},
	pages = {1073--1076},
}

@inproceedings{llminBG,
  title={Adaptive In-Context Learning with Large Language Models for Bundle Generation},
  author={Sun, Zhu and Feng, Kaidong and Yang, Jie and Qu, Xinghua and Fang, Hui and Ong, Yew-Soon and Liu, Wenyuan},
  booktitle={Proceedings of the 47th International ACM SIGIR Conference on Research and Development in Information Retrieval},
  pages={966--976},
  year={2024}
}

@inproceedings{BRUCE,
  title={BRUCE: bundle recommendation using contextualized item embeddings},
  author={Avny Brosh, Tzoof and Livne, Amit and Sar Shalom, Oren and Shapira, Bracha and Last, Mark},
  booktitle={Proceedings of the 16th ACM Conference on Recommender Systems},
  pages={237--245},
  year={2022}
}

@article{ICL,
  title={A survey on in-context learning},
  author={Dong, Qingxiu and Li, Lei and Dai, Damai and Zheng, Ce and Ma, Jingyuan and Li, Rui and Xia, Heming and Xu, Jingjing and Wu, Zhiyong and Liu, Tianyu and others},
  journal={arXiv preprint arXiv:2301.00234},
  year={2022}
}

@inproceedings{ma2024leveraging,
  title={Leveraging multimodal features and item-level user feedback for bundle construction},
  author={Ma, Yunshan and Liu, Xiaohao and Wei, Yinwei and Tao, Zhulin and Wang, Xiang and Chua, Tat-Seng},
  booktitle={Proceedings of the 17th ACM International Conference on Web Search and Data Mining},
  pages={510--519},
  year={2024}
}

@inproceedings{li2023auto,
  title={Auto graph filtering for bundle recommendation},
  author={Li, Xiang-Long and Xi, Wu-Dong and Xing, Xing-Xing and Wang, Chang-Dong},
  booktitle={2023 IEEE International Conference on Data Mining (ICDM)},
  pages={299--308},
  year={2023},
  organization={IEEE}
}

@inproceedings{chen2019matching,
  title={Matching user with item set: Collaborative bundle recommendation with deep attention network.},
  author={Chen, Liang and Liu, Yang and He, Xiangnan and Gao, Lianli and Zheng, Zibin},
  booktitle={IJCAI},
  pages={2095--2101},
  year={2019}
}

@inproceedings{chang2020bundle,
  title={Bundle recommendation with graph convolutional networks},
  author={Chang, Jianxin and Gao, Chen and He, Xiangnan and Jin, Depeng and Li, Yong},
  booktitle={Proceedings of the 43rd international ACM SIGIR conference on Research and development in Information Retrieval},
  pages={1673--1676},
  year={2020}
}

@inproceedings{CLHE,
  title={Leveraging multimodal features and item-level user feedback for bundle construction},
  author={Ma, Yunshan and Liu, Xiaohao and Wei, Yinwei and Tao, Zhulin and Wang, Xiang and Chua, Tat-Seng},
  booktitle={Proceedings of the 17th ACM International Conference on Web Search and Data Mining},
  pages={510--519},
  year={2024}
}

@article{UHBR,
  title={Unifying multi-associations through hypergraph for bundle recommendation},
  author={Yu, Zhouxin and Li, Jintang and Chen, Liang and Zheng, Zibin},
  journal={Knowledge-Based Systems},
  volume={255},
  pages={109755},
  year={2022},
  publisher={Elsevier}
}

@inproceedings{BiLSTM,
  title={Learning fashion compatibility with bidirectional lstms},
  author={Han, Xintong and Wu, Zuxuan and Jiang, Yu-Gang and Davis, Larry S},
  booktitle={Proceedings of the 25th ACM international conference on Multimedia},
  pages={1078--1086},
  year={2017}
}

@inproceedings{MultiVAE,
  title={Variational autoencoders for collaborative filtering},
  author={Liang, Dawen and Krishnan, Rahul G and Hoffman, Matthew D and Jebara, Tony},
  booktitle={Proceedings of the 2018 world wide web conference},
  pages={689--698},
  year={2018}
}

\clearpage
% \appendix
% \thispagestyle{empty}

% Supplementary material: To improve readability, you must use a single-column format for the supplementary material.
% \onecolumn
% \aistatstitle{Supplementary Materials}

% \input{chapters/appendix}

\end{document}